%% file: main.tex
\documentclass[letterpaper, 10 pt, conference]{ieeeconf}
\IEEEoverridecommandlockouts
\usepackage{booktabs}

\input{usepackages.tex}
\input{macros.tex}

\usepackage{titlesec}
\titlespacing*{\subsubsection}{0pt}{1.5ex plus 0.5ex minus .1ex}{1em}

\tcolorboxenvironment{theorem}{
  colback=gray!10,
  colframe=black,
  boxrule=0.8pt,
  arc=1mm,
  left=2mm,
  right=2mm,
  top=1mm,
  bottom=1mm
}

\def\BibTeX{{\rm B\kern-.05em{\sc i\kern-.025em b}\kern-.08em
    T\kern-.1667em\lower.7ex\hbox{E}\kern-.125emX}}
    
\begin{document}

\title{CPED-NCBFs: A Conformal Prediction for Expert Demonstration-based Neural Control Barrier Functions}
% Some Random Title can be changed later

\author{
Sumeadh MS$^{*}$, Kevin Dsouza$^{*}$, Ravi Prakash% <- this % stops a space
\thanks{$^{*}$Equal contribution as co-first authors.}
\thanks{All the authors belong to Center for Cyber-Physical Systems, Indian Institute of Science (IISc), Bengaluru, India.}
}

\maketitle
\begin{abstract}
Among the promising approaches to enforce safety in control systems, learning Control Barrier Functions (CBFs) from expert demonstrations has emerged as an effective strategy. However, a critical challenge remains: verifying that the learned CBFs truly enforce safety across the entire state space. This is especially difficult when CBF is represented using neural networks (NCBFs). Several existing verification techniques attempt to address this problem including SMT-based solvers, mixed-integer programming (MIP), and interval or bound-propagation methods but these approaches often introduce loose, conservative bounds. To overcome these limitations, in this work we use CPED-NCBFs a split-conformal prediction based verification strategy to verify the learned NCBF from the expert demonstrations. 
We further validate our method on point mass systems and unicycle models to demonstrate the effectiveness of the proposed theory.
\end{abstract}

\section{Introduction}
\label{section: Introduction}
\input{1_introduction}

\section{Preliminaries}
\label{section: Background}
\input{2_preliminaries}

\section{Problem Formulation}
\label{section: Problem formulation}
\input{3_problem_formulation}

\section{Methodology}
\label{section: NCBF}
\input{4_conformal_NCBF}

\section{Experiments}
\label{section: experiments}
\input{5_experiments}

\section{Results and Discussion}
\label{section: Results}
\input{6_Results}

\section{Conclusions}
\label{section: Conclusions}
\input{7_conclusion}

\label{section: References}
\input{references}

\end{document}

%% file: usepackages.tex
\usepackage[colorlinks]{hyperref}
\hypersetup{
    colorlinks,
    linkcolor=black,
    citecolor=black,
    filecolor=magenta,
    urlcolor=cyan,
}

\usepackage{resizegather}

\usepackage{cite}
\usepackage{amssymb,amsfonts,mathrsfs}
\usepackage{graphicx}
\usepackage{textcomp}
\usepackage{xcolor}
\usepackage{array}
\usepackage{textcomp}
\usepackage{tablefootnote}
\usepackage{threeparttable}
\usepackage{caption, subcaption}
\usepackage{bbm}
\usepackage{dsfont}
\usepackage{tikz}
\usepackage{tcolorbox}
\usepackage{graphicx}
\usepackage{tkz-euclide}
\usepackage{algorithm}
\usepackage{algpseudocode}
\usetikzlibrary{positioning}
\usetikzlibrary{shapes}
\usetikzlibrary{shapes.misc}
\usetikzlibrary{shapes.geometric}
\usetikzlibrary{plotmarks}
\usetikzlibrary{intersections}
\usetikzlibrary{calc}
\usetikzlibrary{fit}
\usetikzlibrary{patterns,tikzmark}
\usetikzlibrary{matrix,decorations.pathreplacing,calc}

\tikzset{cross/.style={cross out, draw, 
         minimum size=2*(#1-\pgflinewidth), 
         inner sep=0pt, outer sep=0pt}}

%% file: macros.tex
\newcommand{\state}[0]{x}

\newcommand{\vertiii}[1]{{\left\vert\kern-0.25ex\left\vert\kern-0.25ex\left\vert #1 
    \right\vert\kern-0.25ex\right\vert\kern-0.25ex\right\vert}}

\newtheorem{theorem}{Theorem}

\newtheorem{objective}{Objective}
\newtheorem{lemma}{Lemma}

\makeatletter
\renewcommand{\fps@figure}{htp}
\renewcommand{\fps@table}{htp}
\makeatother

\numberwithin{equation}{section}

\newcommand{\RR}{\ensuremath{\mathbb{R}}}
\newcommand{\cbf}{h}

\newcommand{\statedim}{n}
\newcommand{\inputdim}{m}
\newcommand{\xx}{x}

\newcommand{\uu}{u}

\newcommand{\Ss}{\ensuremath{\mathcal{S}}}

\newcommand{\D}{\ensuremath{\mathcal{D}}}

\newcommand{\C}{\ensuremath{\mathcal{C}}}

\newcommand{\N}{\ensuremath{\mathcal{N}}}

\newcommand{\U}{\ensuremath{\mathcal{U}}}
\newcommand{\B}{\ensuremath{\mathcal{B}}}

\DeclareMathOperator*{\bd}{bd}

\DeclareMathOperator*{\inte}{int}

\newcommand{\Zdynamics}{Z_{\mathrm{dyn}}}

%% file: 1_introduction.tex
Modern autonomous systems increasingly operate in dynamic, uncertain environments, where ensuring safety is of high priority. 
Several paradigms have been proposed for safety-critical control, including Safe Reinforcement Learning \cite{achiam2017constrained, NEURIPS2022_9a8eb202, NIPS2017_766ebcd5}, Hamilton–Jacobi reachability~\cite{8263977, tayal2025physics} etc. Among these, Control Barrier Functions (CBFs)\cite{Ames_2017, ames2019control} have emerged as a powerful tool. These functions enable the synthesis of safe controllers by formulating the control problem as a Quadratic Program (QP), which can be solved efficiently in real time using modern solvers. This approach has been successfully employed in a wide range of safety-critical applications, such as adaptive cruise control~\cite{Ames_2017}, aggressive aerial maneuvers~\cite{7525253, tayal2024control}, and legged robot locomotion~\cite{ames2019control, nguyen2015safety,tayal2023safe}. In all these domains, the safety and performance guarantees are inherently tied to the specific CBF used.

To address the increasing complexity of systems and environments, recent work has explored the use of neural networks to parameterize CBFs, resulting in neural control barrier functions (NCBFs) \cite{zhao2020barrier, abate2021fossil, 
tayal2024learning,
robey2020learning, dawson2022robust }. These NCBFs are capable of capturing complex and high-dimensional safe sets that are analytically intractable. However, verifying that a learned NCBF satisfies safety constraints globally, especially outside the support of training data, remains an open problem. Existing verification techniques such as SMT solvers, mixed integer programming (MIP), and interval-bound propagation attempt to offer formal guarantees but introduce overly conservative safety margins, especially in high-dimensional spaces \cite{   tayal2024semi, zhao2022verifying}.

To overcome these limitations, \cite{tayal2025cpncbf} introduced CP-NCBF (Neural CBF Prediction) as a split-conformal prediction framework that transforms learned NCBF into probabilistically verified safety certificates. Rather than relying on symbolic or exhaustive verification, CP-NCBF uses held-out data to calibrate the uncertainty in the learned CBF, producing confidence bounds on its safety performance. This data-driven verification tries to avoid the conservativeness of the worst-case analysis.

Although CP-NCBF offers promising confidence bounds, learning effective CBFs remains sample-inefficient, especially in high-dimensional or safety-critical domains. Naive sampling of state space often wastes resources on irrelevant or already safe regions while
potentially failing to capture the nuanced boundaries near
constraint violations. To address this, expert demonstrations can be used to guide the learning process. Expert trajectories inherently encode safe and task-relevant behaviors \cite{tayal2025genosilgeneralizedoptimalsafe}, enabling imitation-based CBF learning that focuses on meaningful regions close to safety boundaries. Recent studies have shown that demonstrations can serve as implicit supervision for constructing control-theoretic certificates \cite{robey2020learning, dawson2022safe}, particularly when integrated with differentiable programming and end-to-end learning frameworks.

However, while this approach
facilitates learning, it also introduces challenges in verifying
that the learned CBF generalizes beyond the demonstrated
trajectories. This problem becomes especially important when dealing with neural CBFs trained on limited expert data. In this work, our aim is to address this gap by introducing a verification approach tailored to demonstration-driven NCBF learning.

% \mcom{ you may want to fix this sentence, you need to
% explicitly clarify how or why it is faster (maybe somewhere
% in the methodology section), is it because of the intrinsic
% nature of the framework being proposed or whether it is
% empirically claimed through experimentation}
This work makes the following contributions:
\begin{itemize}
    \item We introduce CPED-NCBFs, a conformal prediction-based framework for learning Neural Control Barrier Functions (NCBFs) from expert demonstrations in imitation learning settings, offering probabilistic safety guarantees.
    \item We validate our framework on benchmark dynamical systems including point mass and unicycle models demonstrating strong empirical safety performance.
\end{itemize}
The rest of the sections contain the following. Section II presents preliminaries on stochastic CBFs and safety-critical control. Section III formulates the neural CBF learning problem with expert demonstrations, introducing our key constructs of local safe ($\mathcal{D}$) and unsafe ($\mathcal{N}$) regions. It also includes the Scenario Optimization Problem for margin calibration (SOP-\ref{eq:scp}). Section IV details how we solve the SOP using the conformal prediction-based verification framework and the two-stage training algorithm (Algorithm 1). Section V validates our approach through (i) 3D visualizations of the learned safety limits (Figs.~\ref{fig:cbf-contours}), (ii) trajectory comparisons between FM-NCBF (Fixed Margin) and our CPED-NCBF (Figs.~\ref{fig:pointmass-rollouts},~\ref{fig:trajectories})
, and (iii) quantitative safety rate analysis. Section VI concludes with broader implications and future directions.

% \textcolor{red}{RAVI: Explicitly list the contributions. For example: "This work extends \cite{tayal2025cpncbf} by (1) introducing a framework for learning NCBFs specifically from expert trajectories, defining local safe/unsafe regions (D,N) around these trajectories, and (2) demonstrating how split-conformal prediction can be tailored to verify the generalization of these locally learned NCBFs."}

%% file: 2_preliminaries.tex
Let $\mathbb{R}$ and $\mathbb{R}_{\ge 0}$ be the set of real and non-negative real numbers, respectively, and $\mathbb{R}^n$ the set of $n$-dimensional real vectors. For $\epsilon>0$ and $p\ge 1$, we let $\mathcal{B}_{\epsilon,p}(\bar{x}):=\{x\in\mathbb{R}^n \mid \|x-\bar{x}\|_p\le \epsilon\}$ denote the closed $p$ norm ball around $\bar{x}\in\mathbb{R}^n$. For a given set $S$, we denote by $\bd(S)$, $\text{int}(S)$, and $S^c$ the boundary, interior, and complement of $S$, respectively. For two sets $S_1$ and $S_2$, we denote their Minkowski sum by $S_1\oplus S_2:=\{x_1+x_2\in\mathbb{R}^n \mid x_1\in S_1, x_2\in S_2\}$. A continuous function $\alpha:\mathbb{R}\to\mathbb{R}$ is an extended class function $\mathcal{K}$ if it is strictly increasing with $\alpha(0)=0$. The inner-product between two vectors $x,y\in\mathbb{R}^n$ is denoted by $\langle x,y\rangle$.

\subsection{System Model and Safety Definition}
We consider a control-affine nonlinear dynamical system defined by the state $x \in \mathcal{X} \subseteq \mathbb{R}^{n}$, the control input $u \in \mathcal{U} \subseteq \mathbb{R}^{m}$, and governed by the following dynamics:

\begin{equation}
\dot{x} = f(x) + g(x) u,
\label{eq:dynamics}
\end{equation}

where $f: \mathbb{R}^{n} \rightarrow \mathbb{R}^{n}$ and $g: \mathbb{R}^{n} \rightarrow \mathbb{R}^{n \times m}$ are Lipschitz continuous functions locally. We are given a set $\mathcal{C} \subseteq \mathcal{X}$ that represents the safe states of the system. Furthermore, the system is controlled by a Lipschitz continuous control policy $\pi: \mathbb{R}^{n} \rightarrow \mathbb{R}^{m}$. Our focus lies in ensuring the safety of this dynamical system.

\subsection{Valid control barrier functions}

Consider next a {twice} continuously differentiable function $\cbf:\RR^\statedim\to\RR$, and define the set 
\begin{align}\label{eq:set_C}
\C:=\{\xx\in\RR^\statedim \, \big{|} \, \cbf(\xx)\ge 0\},
\end{align}
as the set that we wish to certify as safe, i.e., the set $\C$ satisfies the prescribed safety specifications and can be made forward invariant through an appropriate choice of control action. We further assume that $\C$ has non-empty interior, and let $\D$ be an open set such that $\D\supset \C$.  The function $\cbf(\xx)$ is said to be a \emph{valid control barrier function} on $\D$ if there exists a locally Lipschitz continuous extended class $\mathcal{K}$ function $\alpha:\RR\to\RR$ such that

\begin{align}\label{eq:cbf_const}
	\begin{split}
		\sup_{\uu\in \U}   %\frac{\partial \cbf(\xx)}{\partial \xx}
		&\underbrace{\langle \nabla \cbf(\xx), f+gu\rangle}_{\substack{\text{Change in } h \text{ along} \\\text{all dynamics}}}+\alpha(\cbf(\xx)) \ge 0
	\end{split}
\end{align} 
holds for all $\xx\in\D$, where $\U\subset\RR^\inputdim$ defines constraints on the control input $u$. 
\subsection{ Safe Controller Synthesis using CBFs}

Quite often, we have a reference control policy, $\pi_{r e f}(x)$, designed to meet the performance requirements of the system. However, such controllers often lack safety guarantees. To ensure the system meets its safety requirements while preserving performance, the reference controller must be minimally adjusted to incorporate safety constraints. This adjustment can be accomplished using the Control Barrier Function-based Quadratic Program (CBF-QP), described as follows:

\begin{align}
\pi_{\text {safe }}(x) = 
& \min _{u \in \mathbb{U} \subseteq \mathbb{R}^{m}}\left\|u - \pi_{\text {ref }}\right\|^{2} \notag \\
\text {s.t. } 
& \mathcal{L}_{f} h(x) + \mathcal{L}_{g} h(x) u + \kappa(h(x)) \geq 0
\label{eq:safe_policy}
\end{align}

The CBF-QP framework facilitates the synthesis of a provably safe control policy, $\pi_{s a f e}(x)$, by leveraging a valid $\mathrm{CBF}, h$, while staying close to the reference controller to preserve system performance.

%% file: 3_problem_formulation.tex
To formalize the previous discussion, we distinguish between geometric safety specifications i.e., those defined over subsets of the state space $x \in \mathbb{R}^n$ and the set $\mathcal{C}$ defined in equation~\eqref{eq:set_C}, certified safe by a Control Barrier Function (CBF). To that end, let $\mathcal{S} \subseteq \mathbb{R}^n$ denote the desired geometric safe set.

To learn a valid CBF, we assume that we are given a set of \emph{expert trajectories} consisting of $N_1$ discretized data-poi      nts $\Zdynamics:=\{(\xx_i,\uu_i)\}_{i=1}^{N_1}$ such that $\xx_i\in\inte(\Ss)$. For $\epsilon>0$, we define the sets
\begin{align}\label{eq:set_D}
    \D':=\bigcup_{i=1}^{N_1}\B_{\epsilon,p}(\xx_i)\qquad\text{and}\quad \ \D :=  \text{int}(D')
\end{align} 
$\mathcal{D}'$ should be thought of as a “layer” of width $\epsilon$ surrounding the sample points $\mathbf{x}_i$. 
$\mathcal{D}$ is the interior of this layer and serves as the effective domain for enforcing the CBF constraints. Conditions on $\epsilon$ will be specified later to ensure the validity of the learned CBF.
For $\sigma>0$, we define the set of  unsafe labeled states 
\begin{align*}
    \N:=\{\bd(\D)\oplus\B_{\sigma,p}(0)\} \setminus \D,
\end{align*} 
which should be thought of as a ``layer'' of width $\sigma$ surrounding the set $\D$.  As will be made clear in the sequel, by enforcing that the value of the learned CBF $\cbf(\xx)$ is negative on the set $\N$, which can be accomplished through appropriate sampling, we ensure that the zero level set $\{ \xx \in \RR^n \, | \, \cbf(\xx) = 0 \}$ is contained within the set $\D$, which is a necessary condition for $\cbf(\xx)$ to be valid.

While the above definition of a CBF is specified over all of $\RR^n$, e.g., the definition of the set $\C$ in equation \eqref{eq:set_C} considers all $x\in \RR^n$ such that $h(x) \geq 0$, we make a minor modification to this definition in order to restrict the domain of interest to the set $\N\cup\D$, i.e., we will certify that $h(x)$ is a valid \emph{local} CBF over the set $\D$ with respect to the set
\begin{align} \label{eq:local_C}
\C:=\{\xx\in\N\cup\D\, \big{|} \, \cbf(\xx)\ge 0\}.
\end{align}

This restriction is natural, as we are learning a CBF $h(x)$ from data sampled only over the domain where the expert data exists.
Let, $h_{\theta}(\state)$, be a Neural CBF (NCBF) parameterized by a deep neural network (DNN) trained via a semi-supervised learning scheme. In the absence of a known CBF, and consequently the true safe set $\mathcal{C}$, we begin with approximate sets constructed from sampled data. Keeping that in mind we formulate the following lemmas which are derived from ~\cite{zhao2020synthesizing}.

\begin{lemma}\label{lem:safe}
Let $h_\theta(\mathbf{x})$ be Lipschitz continuous with local constant $L_{h_\theta}(\mathbf{x})$.
Let $\gamma_s > 0$, and let $X_s$ be an $\epsilon$-net of $\mathcal{D}$ with $\epsilon \le \gamma_s / L_{h_\theta}(x_i)$ for all $x_i \in X_s$. If
\begin{align}
    h_\theta(x_i) \ge \gamma_s
\end{align}
then $h_\theta(x) > 0$ for all $x \in \mathcal{D}$, where
\begin{equation}\label{eq:bXsafe}
\mathcal{X}_s = \left\{ x_i \in \mathcal{D} \,\big|\, \inf_{x \in \mathcal{X}_u} \|x - x_i\|_p \ge \frac{\gamma_s + \gamma_u}{\mathcal{L}_{h_\theta}} \right\}.
\end{equation}
\end{lemma}

\begin{lemma}\label{lem:unsafe}
Let $h_\theta(\mathbf{x})$ be Lipschitz continuous with local constant $L_{h_\theta}(\mathbf{x})$.  
Let $\gamma_u > 0$, and let $X_u$ be an $\bar{\epsilon}$-net of $\mathcal{N}$ with $\bar{\epsilon} < \gamma_u / L_{h_\theta}(x_i)$ for all $x_i \in X_u$. If
\begin{align}
    h_\theta(x_i) \le -\gamma_u
\end{align}
then $h_\theta(x) < 0$ for all $x \in \mathcal{N}$.
\end{lemma}

\begin{lemma}\label{lem:derivative}
Suppose $q(x)$ is Lipschitz continuous with constant $L_q(x)$.  
Let $\gamma_d > 0$, and let $X_d$ be an $\epsilon$-net of $\mathcal{D}$ with $\epsilon\leq \gamma_d/L_q(x_i)$ for all $x_i\in X_d$. If
\begin{align}
    \langle \nabla h_\theta(x_i), f(x_i) + g(x_i) u_i \rangle 
    &\ge -\alpha(h_\theta(x_i)) + \gamma_d
\end{align}
for all $x_i \in X_d$, then $q(x) \ge 0$ for all $x \in \mathcal{D}$.
\end{lemma}

Here, $\gamma$ depends on $\epsilon$ and $L$. If $h_\theta(\mathbf{x})$ satisfies the above propositions, we have that for some non-empty set $\mathcal{C}$ we have $\mathcal{C} \subset \mathcal{D} \subseteq \mathcal{S}$. This implies that there exists some control input $u \in \mathcal{U}$ that renders both the set $\mathcal{C}$, and the set $\mathcal{D}$ forward invariant.

A key challenge in choosing $\gamma$ is its dependence on the Lipschitz constant: higher values of $L$ require larger $\gamma$ to account for steep variations, leading to conservative behavior and overly restrictive safe sets. To mitigate this, we introduce the \textit{Scenario Optimization Problem (SOP)}~\eqref{eq:scp}, for data-driven tuning. The SOP maintains a balance between constraint satisfaction and generalization while providing probabilistic safety guarantees.

\begin{equation}\label{eq:scp}
    \mathrm{SOP}:\begin{cases}
    \boldsymbol{\gamma} = [\gamma_i]_{i \in \{s,u,d\}} \\
    \text{min} \quad \mathbf{1}^\top \boldsymbol{\gamma} \\
    \text{s.t.} \quad q_1(x_i) \leq -\gamma_s, \quad \forall x_i \in \mathcal{X}_s, \\
    \qquad\; q_2(x_i) \leq -\gamma_u, \quad \forall x_i \in \mathcal{X}_u, \\
    \qquad\; q_3(x_i) \leq -\gamma_d, \quad \forall x_i \in \mathcal{X}_d, \\
    \qquad\; \boldsymbol{\gamma} \in \mathbb{R}^3.
    \end{cases}
\end{equation}

where $q_{k}(x), k \in\{1,2,3\}$ are defined as in \eqref{eq:q_conditions}. 

\begin{equation} \label{eq:q_conditions}
    \begin{aligned}
        q_{1}(x)=& \left( -h_{\theta}(x)\right) \mathds{1}_{\mathcal{X}_{s}}, \quad
         q_{2}(x)= \left(h_{\theta}(x)\right) \mathds{1}_{\mathcal{X}_{u}}, \\
         q_{3}(x)=&  -\frac{\partial h_{\theta}}{\partial x} (f(x)+g(x) u(x)) -  \kappa\left(h_{\theta}(x)\right)
         \mathds{1}_{\mathcal{X}_{d}}
    \end{aligned}
\end{equation}

\begin{objective}
\label{obj:custom_sop}
Given a continuous-time control-affine nonlinear dynamical system and the datasets $\mathcal{X}_s$, $\mathcal{X}_u$, and $\mathcal{X}_d$ derived from expert demonstrations, our objective is to develop a framework for synthesizing a Neural Control Barrier Function (NCBF) $h_\theta$ using per-constraint robustness margins $\boldsymbol{\gamma} = [\gamma_s, \gamma_u, \gamma_d]^\top$, such that it satisfies the Scenario Optimization Problem (SOP) in Equation~\eqref{eq:scp} with a user-specified probabilistic confidence level. This ensures that the learned NCBF and the associated controller guarantee safety across the state space with high probability.
\end{objective}

%% file: 4_conformal_NCBF.tex
In this section, we propose an algorithmic approach to address the problem in Section 3. Our approach centers on two key components: loss-based training of the NCBF and a conformal prediction-based estimation of a robustness margin.

\subsubsection{Loss Construction}
The Neural CBF $h_\theta(x)$ is modeled as a feedforward neural network parameterized by $\theta$ to satisfy the CBF constraints. To enforce these constraints, we define a composite loss over three regions: the safe set $\mathcal{X}_s$, the unsafe set $\mathcal{X}_u$, and the derivative-constrained region $\mathcal{X}_d$:
\begin{equation}
L_\theta(\theta) = \lambda_s L_1 + \lambda_u L_2 + \lambda_d L_3,
\label{eq:total_loss}
\end{equation}
where
\begin{align}
L_1(\theta) &= \frac{1}{N} \sum_{x_i \in X_s} \max(0, q_1(x_i)+\gamma_s^*), \label{eq:L1} \\
L_2(\theta) &= \frac{1}{N} \sum_{x_i \in X_u} \max(0, q_2(x_i)+\gamma_u^*), \label{eq:L2} \\
L_3(\theta) &= \frac{1}{N} \sum_{x_i \in X_d} \max(0, q_3(x_i)+\gamma_d^*). \label{eq:L3}
\end{align}

Here, $\lambda_s, \lambda_u, \lambda_d \in \mathbb{R}^+$ weight the contributions of each region, and $\boldsymbol{\gamma}^*$ encodes robustness margins to penalize safety violations.

\subsubsection{Safety Quantification via Conformal Prediction}

To solve the proposed Scenario Optimization Problem (SOP), we leverage a conformal prediction-based strategy that estimates the smallest margin $\boldsymbol{\gamma}^*$ needed to satisfy the constraints with high probability. This margin is critical to balancing safety and conservatism: setting $\boldsymbol{\gamma} < \boldsymbol{\gamma}^*$ can result in safety violations and an incomplete CBF, while choosing $\boldsymbol{\gamma} \gg \boldsymbol{\gamma}^*$ introduces unnecessary conservatism in the control policy. Conformal prediction enables us to compute a probabilistic upper bound on the worst-case constraint violation, using a validation dataset, and thereby yields $\boldsymbol{\gamma}^*$ with formal safety. Theorem \ref{thm:safety_cbf} details the calculation of the safety quantiles $\hat{q}_i$.

\setcounter{theorem}{0}
\begin{theorem}[Safety Quantification of Neural CBF]
\label{thm:safety_cbf}
Consider a continuous-time control-affine system as in Equation~(2.1), and a neural CBF $h_\theta(x)$ parameterized by $\theta$. Let $\{x_j^{(i)}\}_{j=1}^N$ be $N$ i.i.d. samples drawn from each constraint set $\mathcal{X}_i$, where $i \in \{s, u, d\}$. Let m be the number of constraint sets.

For each set $\mathcal{X}_i$, compute the conformal scores:
\[
s_j^{(i)} = q_i(x_j^{(i)}),
\]
and let $\hat{q}_i$ be the $(1 - \alpha/m)$ quantile of $\{s_j^{(i)}\}_{j=1}^N$:
\[
\hat{q}_i := \text{Quantile}_{1 - \alpha/m} \left( \{ s_j^{(i)} \}_{j=1}^N \right).
\]

Choose violation level $\epsilon \in (0,1)$ and confidence level $\beta \in (0,1)$ such that:
\[
\mathcal{I}_{1 - \epsilon}(N - l + 1, l) \leq \beta, \quad \text{where } l = \lfloor (N + 1)\alpha/m \rfloor,
\]
and $\mathcal{I}$ denotes the regularized incomplete beta function. Then, with probability at least $1 - \beta$, the following holds simultaneously for each constraint set:
\[
\mathbb{P}_{x \in \mathcal{X}_i} \left( q_i(x) \leq \hat{q}_i \right) \geq 1 - \epsilon \quad \forall i \in \{s, u, d\}.
\]
\end{theorem}

For sufficiently high confidence level $(1 - \beta)$ and safety level $(1 - \epsilon)$, the quantiles $\hat{q}_i$ provide a probabilistic upper bound on constraint violation within each region. A positive $\hat{q}_i$ indicates violations within set $\mathcal{X}_i$, and its magnitude quantifies the worst-case severity in that region.

\begin{algorithm}[t]
\caption{Training Robust Neural CBF with Probabilistic Safety Assurance}
\label{alg:training_combined}
\begin{algorithmic}[1]
\Require Data sets: $\mathcal{X}_s, \mathcal{X}_u, \mathcal{X}_d$; Dynamics: $f, g$; $N$, $\alpha$
\State $x_i \gets \text{Sample}(\mathcal{X}_s, \mathcal{X}_u, \mathcal{X}_d)$
\For{$k \in \{s, u, d\}$}
    \State $\gamma_k \gets 0$
\EndFor
\While{$\mathcal{L}_\theta(\theta) > 0$}
    \State $\mathcal{L}_\theta(\theta) \gets \text{ComputeLoss}(f, g, x_i, \{\gamma_k\})$
    \State $\theta \gets \text{Learn}(\mathcal{L}_\theta, \theta)$
\EndWhile
\For{$k \in \{s, u, d\}$}
    \State Sample $N$ IID states $D_k$ from $\mathcal{X}_k$
    \State Compute $S_i^k = s_i^k(0, D_k)$ for all $i$
    \State Sort $S_i^k$ decreasingly; set $l = \lfloor (N + 1)\alpha/3  \rfloor$
    \State $\hat{q}_k \gets S_l^k$
    \State $\gamma_k \gets \hat{q}_k$
\EndFor
\While{$\mathcal{L}_\theta(\theta) > 0$}
    \State $\mathcal{L}_\theta(\theta) \gets \text{ComputeLoss}(f, g, x_i, \{\gamma_k\})$
    \State $\theta \gets \text{Learn}(\mathcal{L}_\theta, \theta)$
\EndWhile
\State \Return $h_\theta$
\end{algorithmic}
\end{algorithm}

\subsubsection{Training Algorithm}

Training is conducted in two stages. Initially, the Neural Control Barrier Function (NCBF) is trained with margins $\gamma_k = 0$ for each $k \in \{s, u, d\}$. This corresponds to enforcing safety constraints without any slack.

Next, for each set $\mathcal{X}_k$, a conformal score $\hat{q}_k$ is computed using a validation dataset that was strictly excluded from the training process. This separation ensures that the estimated margins reflect the model’s generalization ability rather than overfitting to the training data.
. If any $\hat{q}_k > 0$, indicating potential safety violations, the corresponding margin is updated as $\gamma_k = \hat{q}_k$, and the NCBF is retrained using the updated margins. This process repeats until all $\hat{q}_k \leq 0$ or a maximum number of iterations is reached.

The final safe control policy is obtained by solving a Quadratic Program (QP) that filters a nominal control input $u_{\text{ref}}(x)$ to produce a safe input $u$, subject to the learned NCBF constraints. This approach ensures high-probability satisfaction of safety constraints.

%% file: 5_experiments.tex
We evaluate the effectiveness of our proposed framework on two safety-critical control systems that we implemented and tested: (i)Point mass system  (ii) Unicycle model.

\subsection{Point Mass Collision Avoidance}  

In this section, we present simulation setup for a point mass system. We consider a two-dimensional nonlinear point-mass system defined by the dynamics $\dot{x}_1 = -x_1 + (x_1^2 + \delta)u_1$ and $\dot{x}_2 = -x_2 + (x_2^2 + \delta)u_2$, where $\delta > 0$ is a fixed parameter (set to $\delta = 1$ in our experiments). The state and control vectors are $x = [x,\ y]^\top$ and $u = [v_x,\ v_y]^\top$, respectively. The system is required to remain within a safety region $\mathcal{S} = \{ x : x_1 \leq 1,\ x_2 \leq 1 \}$, which is encoded via a candidate control barrier function (CBF) defined as $h(x) = \min \{1 - x_1,\ 1 - x_2\}$.The goal of the point mass system is to follow the given reference trajectory while staying inside the safe zone.

% \begin{figure}[H]
%     \centering
%     \includegraphics[width=0.5\linewidth]{images/samples.png}
%     \caption{Illustrative plot of the sampling strategy. Points from the safe set $\mathcal{X}_s$ and expert demonstrations $\mathcal{X}_d$ are shown in green, unsafe states from $\mathcal{X}_u$ are shown in red, and expert control-only data $\mathcal{N}$ are shown in black.
%  }
%     \label{fig:sampling}
% \end{figure}   

\subsection{Unicycle Model}

In this section, we present simulation setup for a unicycle model. This setup is inspired by~\cite{zhao2020synthesizing}, where we adapt the double airplane collision avoidance problem into a single-agent formulation. The system dynamics are given by $\dot{\mathbf{x}}(t) = [v(t) \cos \theta(t),\ v(t) \sin \theta(t),\ \omega(t)]^\top$, where the state and control vectors are $\mathbf{x} = [x,\ y,\ \theta]^\top$ and $\mathbf{u} = [v,\ \omega]^\top$, respectively. Expert demonstrations are generated using a model predictive controller (MPC) combined with a pre-trained barrier function defined as $h(\mathbf{x}) = (x - R \sin \theta)^2 + (y + R \cos \theta)^2 + 2R^2 - D_s^2$, refer  \cite{squires2023learning} for more explanation.The goal of the unicycle model is to reach the goal point (5,0,$\pi$) or (-5,0,-$\pi$) while staying inside the safe set.

\begin{figure}[H]
    \centering
    \begin{subfigure}[b]{0.48\linewidth}
        \centering
        \includegraphics[width=\linewidth]{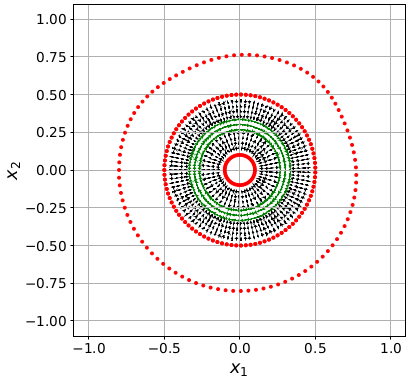}
        \caption{Sampling (Point Mass)}
        \label{fig:sampling1}
    \end{subfigure}
    \hfill
    \begin{subfigure}[b]{0.48\linewidth}
        \centering
        \includegraphics[width=\linewidth]{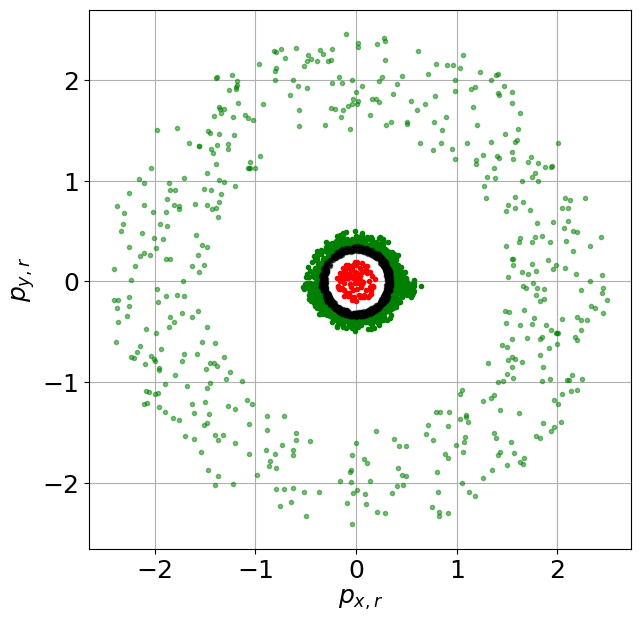}
        \caption{Sampling (Unicycle Model)}
        \label{fig:sampling2}
    \end{subfigure}
    \caption{ Points from the safe set $\mathcal{X}_s$ and expert demonstrations $\mathcal{X}_d$ are shown in green, unsafe states from $\mathcal{X}_u$ in red, and buffer data in black.}
    \label{fig:sampling}
\end{figure}

\vspace{1em}
\textit{Data Generation Strategies}: We simulate trajectories using the system dynamics (point mass or unicycle)  and obtain $\Zdynamics$ which is used to sample data-points. Then, we use boundary-focused radial sampling, with dense sampling near decision boundaries to enforce the proposition \ref{lem:derivative}. The expert samples are drawn radially due to the system's radial symmetry and to reduce the number of samples compared to uniform sampling. The safe set $\mathcal{X}_s$ and the unsafe set $\mathcal{X}_u$ are radially sampled to enforce constraints (\ref{lem:safe},\ref{lem:unsafe}). The expert set \( \mathcal{X}_d \) includes trajectories near critical boundaries. Additionally, a buffer data is generated in a radial band of width \( w \) around the unsafe-safe transition region with controls directed inward or outward to push the agent back to safety. All sets are split into training and test subsets, allowing us to estimate the \( \hat{q} \)-values. An illustration is provided in Figure \ref{fig:sampling}

%% file: 6_Results.tex
\subsection{3D Surface Plot of the Barrier Function}

\begin{figure}
    \centering
    \begin{subfigure}[t]{0.23\textwidth}
        \centering
        \includegraphics[width=\linewidth]{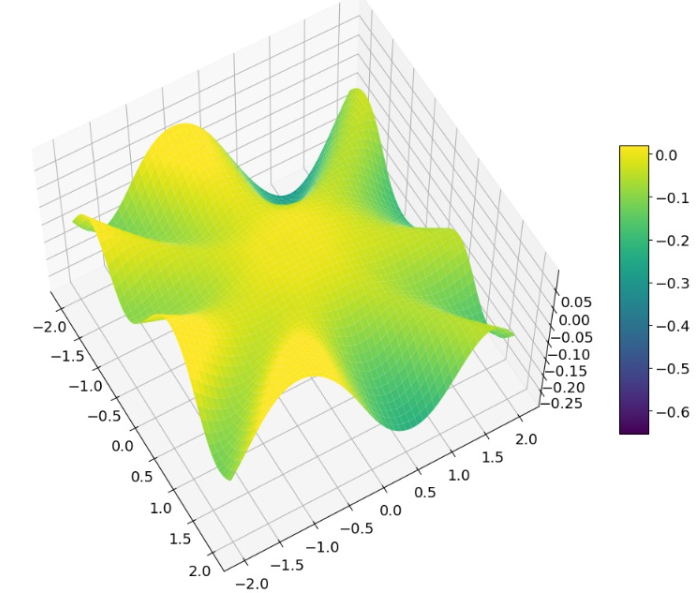}
        \caption{FM-NCBF (Point Mass)}
        \label{fig:cbf-contour-standard-1}
    \end{subfigure}
    \hfill
    \begin{subfigure}[t]{0.23\textwidth}
        \centering
        \includegraphics[width=\linewidth]{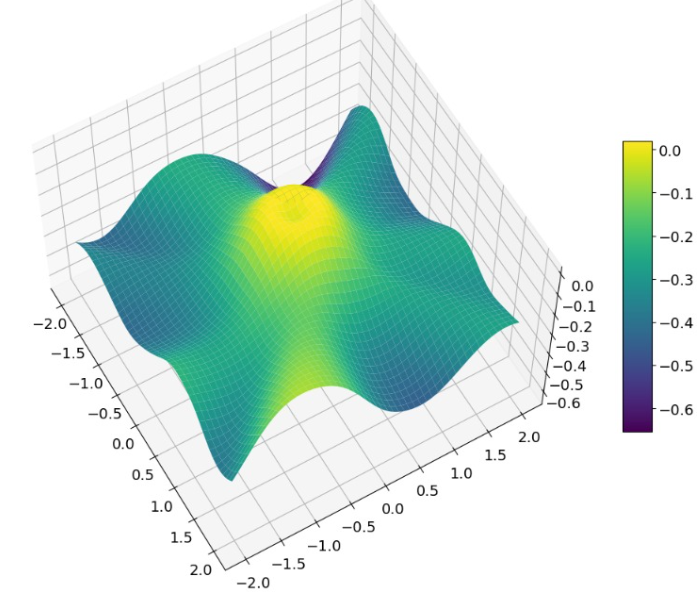}
        \caption{CPED-NCBF (Point Mass)}
        \label{fig:cbf-contour-cp-1}
    \end{subfigure}
    \hfill
    \begin{subfigure}[t]{0.23\textwidth}
        \centering
        \includegraphics[width=\linewidth]{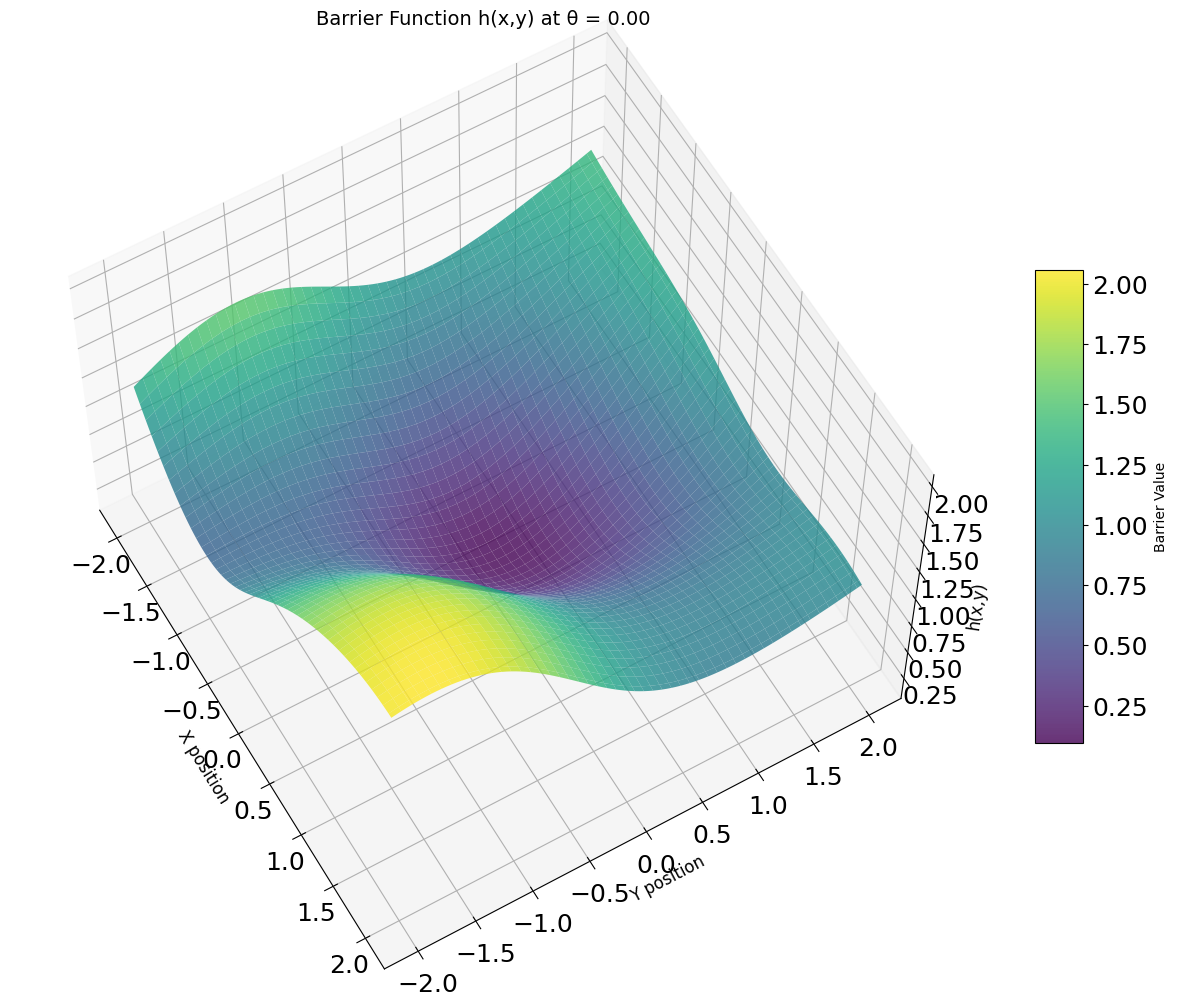}
        \caption{FM-NCBF (Unicycle)}
        \label{fig:cbf-contour-standard-2}
    \end{subfigure}
    \hfill
    \begin{subfigure}[t]{0.23\textwidth}
        \centering
        \includegraphics[width=\linewidth]{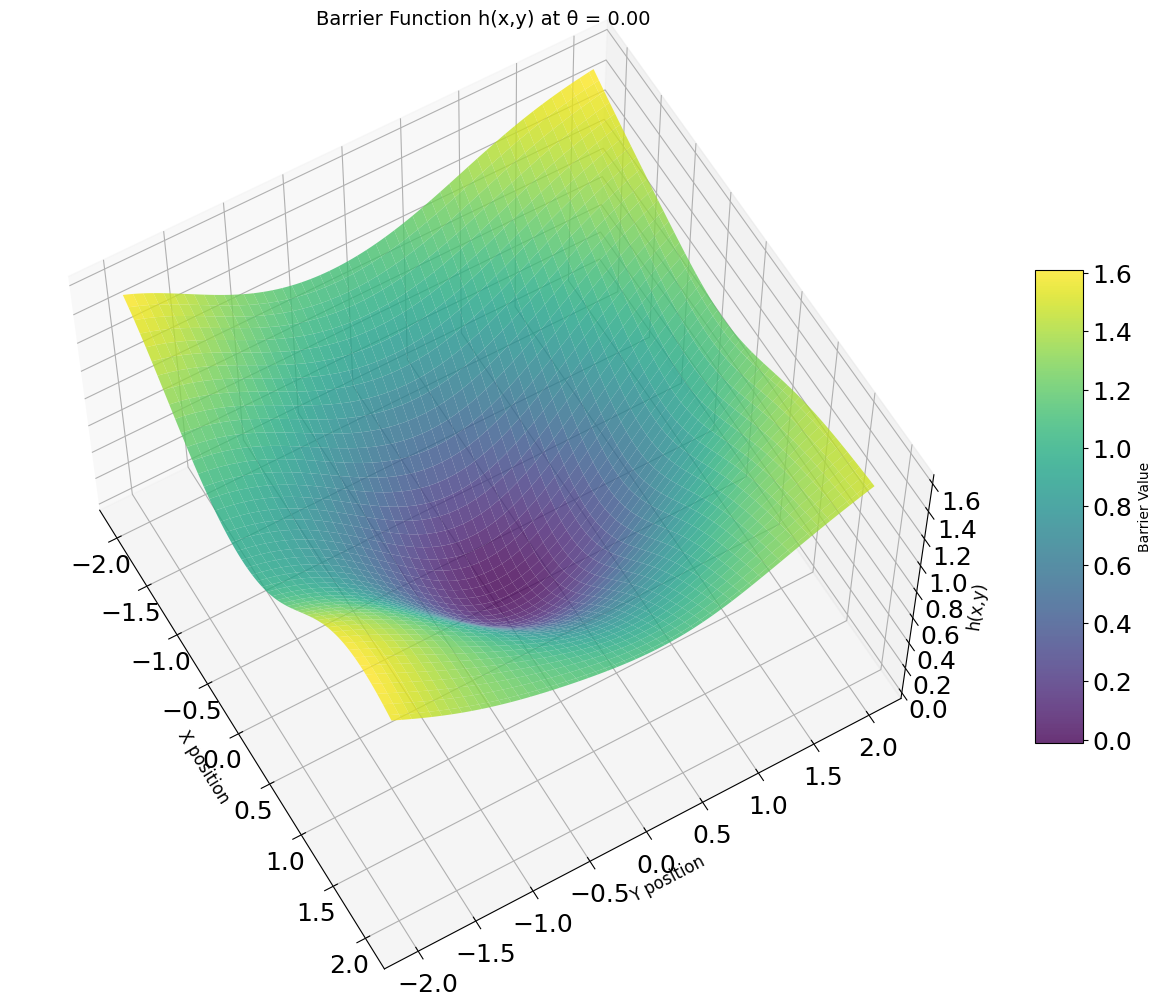}
        \caption{CPED-NCBF (Unicycle)}
        \label{fig:cbf-contour-cp-2}
    \end{subfigure}
    \caption{Comparative 3D Surface Plots}
    \label{fig:cbf-contours}
\end{figure}

Figure~\ref{fig:cbf-contours} shows 3D surface plots of the learned control barrier functions (CBFs) for two systems, each with FM-NCBF (Fixed Margin) and CPED-NCBF variants. Both cases aim to approximate the safety boundaries, the boundary generated by FM-NCBF exhibits a rougher surface with less distinct separation between safe and unsafe regions. In contrast,the CPED-NCBF push safe states deeper into the safe region and unsafe states further into the unsafe domain. This leads to a clearer and more reliable separation, improving safety assurance during control.

\subsection{Trajectory Analysis: CPED-NCBF vs FM-NCBF}

\begin{figure}[h]
    \centering
    \includegraphics[width=0.45\textwidth]{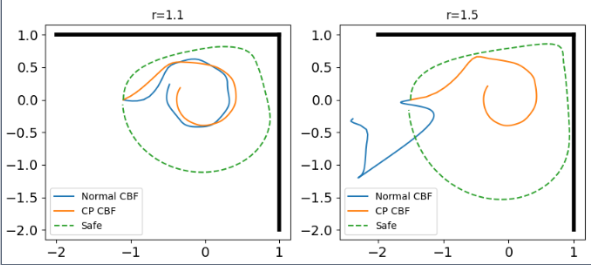}
    \caption{Rollout trajectory using FM-NCBF. Green dots denote the safety margin.}
    \label{fig:pointmass-rollouts}
\end{figure}

 Figure~\ref{fig:pointmass-rollouts} represents the trajectories for the point mass system. With the FM-NCBF, rollout trajectories occasionally violate the defined safety constraints, particularly when initial states are close to the boundary. This highlights the limitations of the model’s decision boundary.

\begin{figure}[H]
    \centering
    \begin{subfigure}[t]{0.23\textwidth}
        \centering
        \includegraphics[width=\linewidth]{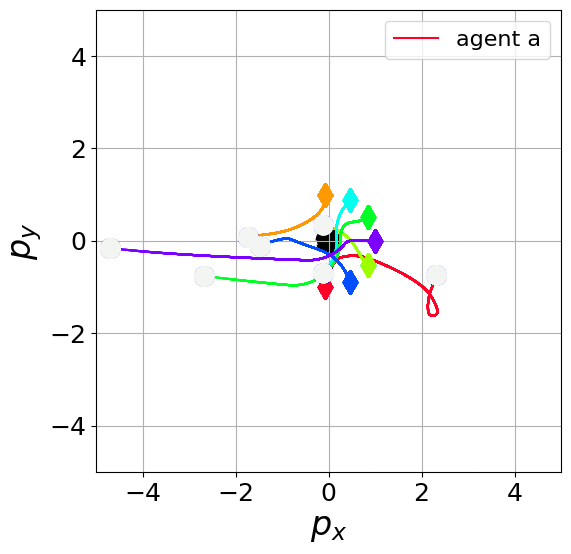}
        \caption{FM-NCBF}
        \label{fig:trajectories_without_cp}
    \end{subfigure}
    \hfill
    \begin{subfigure}[t]{0.23\textwidth}
        \centering
        \includegraphics[width=\linewidth]{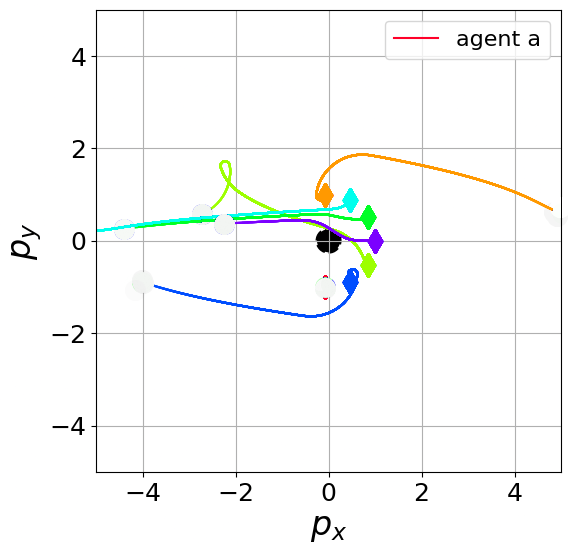}
        \caption{CPED-NCBF}
        \label{fig:trajectories_after_cp}
    \end{subfigure}
    \caption{Trajectory plots comparing FM-NCBF and CPED-NCBF. The diamonds indicate the start point and white dots indicate the end point of the trajectory. The black circle in the center indicates the barrier.}
    \label{fig:trajectories}
\end{figure}

 From Figure~\ref{fig:trajectories}, we can further confirm the behavior shown by the unicycle model in the previous section. The FM-NCBF in Figure~\ref{fig:trajectories_without_cp} attempts to improve performance at the expense of safety, with some trajectories even attempting to cross the barrier. In contrast, the CPED-NCBF effectively increases the robustness margin, enabling the model to maintain safe behavior while avoiding the barrier more reliably.

\subsection{Quantitative Analysis: CPED-NCBF vs FM-NCBF}

We evaluate how well the point mass system generalize to increasing trajectory radii \( r \in [0.1, 3.1] \) under varying training sample sizes. A rollout is marked unsafe if it exits the defined safe region. As shown in Table~\ref{tab:safety_results}, FM-NCBF fails at most radii with few samples and only becomes reliable for \( r \leq 1.1 \) as samples increase. In contrast, CPED-NCBF already performs safely up to \( r = 1.1 \) with 390 samples and extends to \( r \geq 2.5 \) as data grows. 
We do not report safety rates for the point mass system, as rollout performance across increasing radii better captures generalization. Since the goal is to assess robustness to spatial deviation, this metric is more meaningful here.

\begin{table}[h]
\centering
\caption{Generalization radius vs. dataset size.}
\label{tab:safety_results}
\begin{tabular}{>{\raggedright}p{2cm}cc}
\toprule
\textbf{Samples} & \textbf{FM-NCBF} & \textbf{CPED-NCBF} \\
\midrule
390   & Unsafe at all \( r \)      & Safe up to \( r = 1.1 \) \\
650   & Safe up to \( r = 0.7 \)    & Safe up to \( r = 2.5 \) \\
910   & Safe up to \( r = 1.1 \)    & Safe up to \( r = 2.7 \) \\
1430  & Safe up to \( r = 1.1 \)    & Safe up to \( r = 2.7 \) \\
\bottomrule
\end{tabular}
\end{table}

Subsequently, we simulate 100 trajectories and compare the safety rates for the unicycle model. We tabulate the results with respect to the change in the total number of samples. We notice here that the FM-NCBF slightly outperforms the CPED-NCBF with large datasets, suggesting an exact approximation in the barrier function. However, the CPED-NCBF sigificantly outperforms the FM-NCBF when the availability of samples is fewer in number. In this case, the $\gamma$ value becomes high, making the barrier highly conservative.
\begin{table}[h]
\centering
\caption{Safety rate (\%) comparison}
\label{tab:safety_comparison}
\begin{tabular}{>{\raggedright}p{2cm}cc}
\toprule
\textbf{Samples} & \textbf{FM-NCBF (\%)} & \textbf{CPED-NCBF (\%)} \\
\midrule
1,000  & 90.5  & 98.8 \\
5,000  & 98.1  & 99.1 \\
10,000 & 99.7  & 99.4 \\
\bottomrule
\end{tabular}
\end{table}

%% file: 7_conclusion.tex
In this paper, we address the challenge of efficiently verifying neural network parameterized control barrier functions (NCBFs) for safety-critical systems. We proposed CPED-NCBF, a split-conformal prediction framework that provides probabilistic safety guarantees for neural CBFs. Through experiments on point-mass systems and unicycle models, we demonstrated that our method achieves superior safety rates compared to existing approaches. This work offers a practical direction for scaling safety verification to complex, high-dimensional systems. Future work will focus on extending our framework to real-world robotic platforms.

%% file: references.tex
\bibliographystyle{IEEEtran}
\bibliography{references.bib}